\documentclass{article}
\usepackage{iclr2024_conference,times}

\usepackage{amsmath,amsfonts,bm}

\def\eqref#1{equation~\ref{#1}}

\def\1{\bm{1}}

\DeclareMathAlphabet{\mathsfit}{\encodingdefault}{\sfdefault}{m}{sl}
\SetMathAlphabet{\mathsfit}{bold}{\encodingdefault}{\sfdefault}{bx}{n}

\usepackage{hyperref}
\usepackage{url}
\usepackage{soul}

\usepackage{multirow}
\usepackage{booktabs}
\usepackage{comment}
\usepackage{mathtools}
\usepackage{tcolorbox}
\usepackage{adjustbox}
\usepackage{xcolor}
\definecolor{darkgreen}{rgb}{0.0, 0.5, 0.0}

\tcbuselibrary{most}

\definecolor{cadmiumgreen}{rgb}{0.0, 0.42, 0.24}

\newcommand{\puser}{p}

\newcommand{\concat}{\mathbin\Vert}
\newcommand{\Dprompt}{\mathcal{D}}
\newcommand{\hproxy}{h_\text{prox}}
\newcommand{\hr}{\ell_\text{hal}}
\newcommand{\lossref}{\ell_\text{ref}}
\newcommand{\lossopt}{\ell_\text{opt}}
\newcommand{\tausim}{\tau_\text{syn}}
\newcommand{\ours}{\textsc{SynTra}}
\newcommand{\thetaog}{\theta_\text{og}}
\newcommand{\phiog}{\phi_\text{og}}
\newcommand{\Dref}{\mathcal{D}_\text{ref}}
\newcommand{\sys}{s}

\newcommand{\lm}{LLM}
\newcommand{\upbetter}{{\textcolor{darkgreen}{\mathbf{(\Uparrow)}}}}
\newcommand{\downbetter}{{\textcolor{darkgreen}{\mathbf{(\Downarrow)}}}}

\newtcolorbox{userinput}{
    colback=gray!10,
    colframe=gray!40,
    fonttitle=\bfseries,
    coltitle=black,
    colbacktitle=gray!60,
    enhanced,
    drop shadow=black!5!white,
    left=8mm,
    right=8mm,
    top=3mm,
    bottom=3mm,
    boxsep=0mm,
    sharp corners=south,
    rounded corners=north,
    title=Prompt:
    
    }

\newcommand\refeqn[1]{(\ref{eqn:#1})}

\newcommand\refsec[1]{Section~\ref{sec:#1}}

\newcommand\reffig[1]{Figure~\ref{fig:#1}}

\newcommand\reftab[1]{Table~\ref{tab:#1}}
\newcommand\refapp[1]{Appendix~\ref{sec:#1}}

\title{Teaching Language Models to Hallucinate Less with Synthetic Tasks}

\author{
Erik Jones,\textsuperscript{\textnormal{1}}\footnotemark[3] ~
Hamid Palangi,\textsuperscript{\textnormal{2}} ~
Clarisse Simões,\textsuperscript{\textnormal{2}} ~
Varun Chandrasekaran,\textsuperscript{\textnormal{3}}\footnotemark[3]  ~
\\
\textbf{
Subhabrata Mukherjee,\textsuperscript{\textnormal{4}}\footnotemark[3] ~
Arindam Mitra,\textsuperscript{\textnormal{2}} ~
Ahmed Awadallah,\textsuperscript{\textnormal{2}} ~
Ece Kamar\textsuperscript{\textnormal{2}}
}
\\[0.5ex]
\hspace{20mm}\textsuperscript{1}\,UC Berkeley~
\textsuperscript{2}\,Microsoft Research\,
\textsuperscript{3}\,UIUC
\textsuperscript{4}\,Hippocratic AI
\phantom{\footnotemark[2]}
\vspace{-3mm}
}

\renewcommand{\paragraph}{\textbf}

\iclrfinalcopy 
\begin{document}

\maketitle
\begin{abstract}
Large language models (LLMs) frequently hallucinate on abstractive summarization tasks such as document-based question-answering, meeting summarization, and clinical report generation, even though all necessary information is included in context. 
However, optimizing LLMs to hallucinate less on these tasks is challenging, as hallucination is hard to efficiently evaluate at each optimization step. 
In this work, we show that reducing hallucination on a \emph{synthetic task} can also reduce hallucination on real-world downstream tasks. 
Our method, \ours{}, first designs a synthetic task where hallucinations are easy to elicit and measure. 
It next optimizes the LLM's system message via prefix-tuning on the synthetic task, and finally transfers the system message to realistic, hard-to-optimize tasks. 
Across three realistic abstractive summarization tasks, \ours{} reduces hallucination for two 13B-parameter LLMs using only a synthetic retrieval task for supervision. 
We also find that optimizing the system message rather than the model weights can be critical; fine-tuning the entire model on the synthetic task can counterintuitively \emph{increase} hallucination. 
Overall, \ours{} demonstrates that the extra flexibility of working with synthetic data can help mitigate undesired behaviors in practice.
\end{abstract}

\renewcommand{\thefootnote}{\fnsymbol{footnote}}
\footnotetext[2]{Correspondence to \texttt{erjones@berkeley.edu} and \texttt{hpalangi@microsoft.com}}
\footnotetext[3]{Work done at Microsoft Research.}
\renewcommand{\thefootnote}{\arabic{footnote}}
\section{Introduction}
\label{sec:intro}
Large language models (LLMs) are prone to hallucinate---i.e., fabricate entities, details, or other content---when generating responses to queries, even when all salient information is included in context. For example, LLMs can make up citations \citep{liu2023evaluating}, add titles when generating biographies \citep{min2023factscore}, or invent new product attributes when advertising \citep{koto2022pretrained}. 
To confidently deploy LLMs, we need methods to monitor and reduce these hallucinations. 

Unfortunately, directly reducing LLM hallucinations on real-world tasks is challenging, in part because we cannot scalably evaluate hallucination during optimization. 
To exhibit this evaluation challenge, suppose the LLM generates the fictional noun ``fixed liability response'' when summarizing a meeting. 
Cheap-to-run rule-based manual verifiers would struggle to (i) decide that this term is worth checking, and (ii) identify whether the term is real or fabricated. 
On the other hand, using humans or LLMs to detect hallucination in long-form outputs is slow, expensive, and error-prone~\citep{guerriero2023looking}. 
It is thus difficult to directly optimize against LLM hallucinations with gradient descent or reinforcement learning, as we cannot efficiently evaluate the loss or reward.

\begin{figure}[t]
  \centering
  \includegraphics[width=0.99\linewidth]{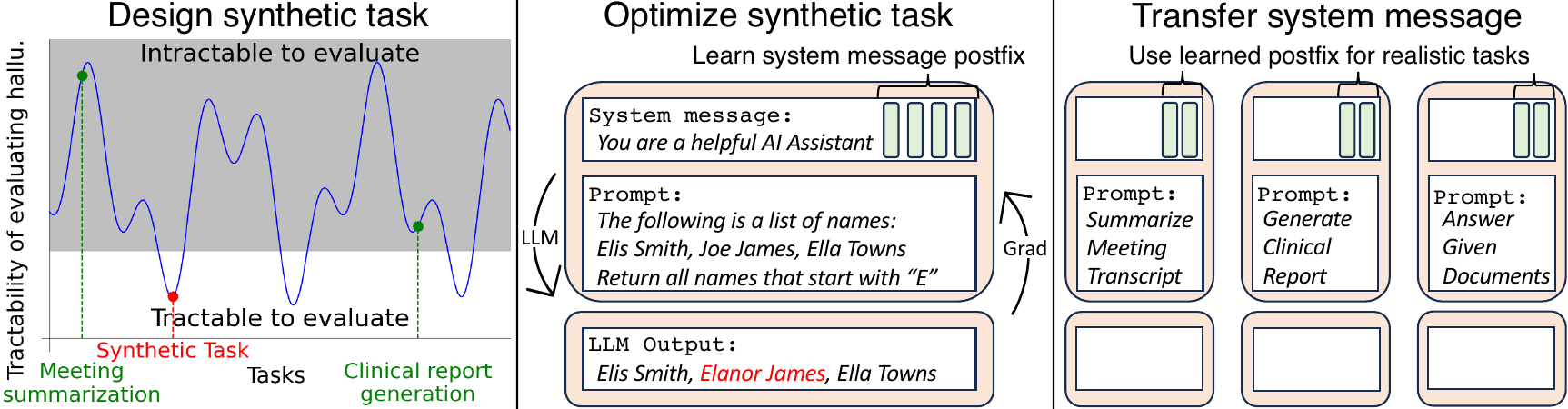}
  \caption{Overview of the \ours{} framework. We first define a synthetic task where hallucination is easy to tractably evaluate. 
  Next, we optimize the LLM system message on this task by learning a continuous postfix via prefix-tuning. We then transfer the learned system message across real tasks.}
  \label{fig:f1} 
\end{figure}

In response, we introduce \ours{}, a method that uses synthetic data to reduce LLM hallucinations (\reffig{f1}). 
\ours{} first designs a \emph{synthetic task} where hallucination can be efficiently and tractably evaluated. It then exploits this tractability by optimizing the LLM system message on the synthetic task via prefix-tuning \citep{li2021prefix}, and finally transfers the system message to realistic tasks. 

The core component of \ours{} is the design of the synthetic task, which provides the only direct supervision signal that captures hallucination. 
For this supervision signal to be sufficient, we argue that the synthetic task must at least satisfy two properties: (i) LLMs should hallucinate frequently on the task, and (ii) hallucination can be cheaply and automatically evaluated on the task. 
The former ensures that optimizing on the synthetic task teaches the model to hallucinate less, while the latter makes optimization tractable. 
Throughout this work, we use the \emph{names retrieval task} as the synthetic task: given a list of random names, we prompt the LLM to retrieve the first $n$ names on the list that start with some letter. We say the model hallucinates if it generates a name that is not on the list. 

\ours{} then optimizes the LLM to reduce hallucination on the synthetic task. 
To do so, we optimize the LLM system message (e.g.,``\emph{You are a helpful AI assistant}'') by appending a continuous postfix to it, then optimizing the postfix. 
We optimize the system message rather than the whole model to learn high-level instructions for how to hallucinate less, 
which we expect to transfer well. 
We additionally optimize the LLM to keep its output constant on a set of reference prompts, so the LLM does not latch onto spurious attributes of the synthetic task. 

We find that \ours{} consistently reduces hallucination across models, tasks, and metrics. 
We evaluate Vicuna v1.1 \citep{weilin2023vicuna} and Orca \citep{mukherjee2023orca} on three realistic tasks: search-and-retrieve, meeting summarization, and clinical report generation. Following, \citet{yue2023automatic}, we measure hallucination with GPT-4, and find that optimizing the system message consistently reduces hallucination: on Orca, \ours{} reduces the hallucination rate by over 7 points on average and 16 points on specific tasks. 
In contrast, fine-tuning the whole model can \emph{increase} the hallucination rate. 
We also test whether \ours{} reduces hallucination according to other metrics, and find its outputs (i) overlap more with references and (ii) contain fewer ungrounded entities.

\ours{} is not without limitations; it requires designing a synthetic task, and reduces hallucination on some models more than others. Nevertheless our work provides encouraging evidence that we can use synthetic data to isolate and optimize against undesired LLM behaviors. 
\section{The \ours{} pipeline}
We describe \ours{} (Synthetic Transfer), our method to reduce hallucination on abstractive summarization tasks using synthetic data. We outline the setup (\refsec{setup}), introduce synthetic tasks (\refsec{simulation-tasks}), then describe how \ours{} reduces hallucination on realistic tasks (\refsec{transfer-to-general}).

\subsection{Setup}
\label{sec:setup}
We study LLMs that take prompts (e.g., ``\emph{Write a clinical report about the following dialog...}'') and produce long-form outputs (e.g., ``\emph{Overview: Patient has...}''). 
The behavior of the LLM is modulated by two features: a system message that provides high-level instructions (e.g., ``\emph{You are a helpful and honest AI assistant})'', and the raw model weights.  
We thus represent the LLM as $\lm_{\phi, \theta}$, a composition of the \emph{system function} $\sys_\phi$ that appends the prompt to the system message $\phi$,\footnote{While the system prompt $\phi$ is usually text, this is challenging to optimize. 
Following~\citet{li2021prefix} we append continuous embeddings to the text, which can be optimized via gradient descent.} 
and the \emph{raw LLM} $f_\theta$ that generates text based on weights $\theta$. Formally, $\lm_{\phi, \theta} = f_\theta \circ \sys_\phi$. 
This decomposition into system messages and prompts is standard for many instruction-tuned LLMs~\citep{openai2023gpt4}. 

We measure hallucination on \emph{abstractive summarization tasks}, where models respond to queries given some context. 
We model an abstractive summarization task $\tau$ (e.g., generate clinical reports from dialogues) as a distribution over prompts $\Dprompt_\tau$, where each prompt $\puser$ can be decomposed into a query $q$ that is appended to context $c$, i.e., $\puser = c \concat q$. For example, the context $c$ might be a dialog between a patient and a doctor, and the query might be ``\emph{Write a clinical report containing ...}''. We assume that the context $c$ always has all required information to correctly respond to query $q$. 

Hallucinations in the abstractive summarization setting are \emph{grounding errors}: errors where the output contains some information that is not supported by the context. 
We define the hallucination function $h$, which takes in a prompt and output and returns 1 if the output is supported by the prompt, and 0 otherwise.
The hallucination rate $\hr$ on task $\tau$ for model parameters $\theta$ and system prompt $\phi$ is
\begin{align}
    \hr(\phi, \theta; \tau) \coloneqq \mathbb{E}_{\puser \sim \Dprompt_\tau} \; h(\puser, \lm_{\phi, \theta}(\puser)),
\end{align}
where the expectation is taken over the distribution $\Dprompt_\tau$ and sampling during LLM decoding. 

The hallucination function $h$ is at best expensive to evaluate for many different user prompts and outputs, and at worse intractable. 
This is especially true for abstractive summarization, as even humans can struggle to read and remember long contexts while evaluating generated summaries \citep{guerriero2023looking}. 
To measure hallucination in general, we rely on proxy metrics $\hproxy$ that can use reference outputs or external databases to approximate $h$. 
Existing work has considered many possible proxy metrics including comparing to ground truth outputs~\citep{nan2021entity}, decomposing outputs into atomic facts and evaluating them separately~\citep{min2023factscore}, using entailment models~\citep{roit2023factually}, or using LLM judges~\citep{yue2023automatic, gao2023rarr}. 

\subsection{Synthetic tasks}
\label{sec:simulation-tasks}
A natural approach to reduce hallucination is to optimize against the hallucination rate $\hr$ over all tasks simultaneously, but this is intractable; the hallucination function $h$ is expensive to evaluate at scale, which rules out direct optimization. 
Optimizing against proxy metrics $\hproxy$ is also problematic, as they may (i) unfaithfully capture hallucination and (ii) still be expensive to optimize. 

Rather than trying to optimize a proxy in general, we instead optimize the true hallucination rate exactly on a carefully constructed \emph{synthetic task}. 
Concretely, we construct a synthetic task $\tausim$ such that the hallucination function restricted to this task, $h\big|_{\Dprompt_{\tausim}}$, 
is easy to automatically evaluate and optimize. 
For example, suppose the prompts contain random first and last names (context) and ask the LLM to retrieve $n$ names that start with some letter (query). 
We can test if the LLM hallucinates by checking whether the each name it generates is in the prompt, and optimize to reduce the corresponding hallucination rate directly.\footnote{In this case, we add constraints until there is a unique non-hallucinated output (e.g., names are outputted in order as a comma-separated list) to avoid averaging over the combinatorially many non-hallucinated outputs. We could efficiently optimize other definitions (e.g., is any name not on the list) with zeroth order methods.}

We seek synthetic tasks $\tausim$ that satisfy two desiderata: (i) the LLM hallucinates frequently on the task, and (ii) we can test for hallucination on the task efficiently. The former means optimizing on the synthetic task teaches the model to hallucinate less, while the latter makes optimization tractable. 

\subsection{Reducing hallucination with \ours{}}
\label{sec:transfer-to-general}
We now present the entire \ours{} pipeline. \ours{} defines a synthetic task (\refsec{simulation-tasks}), optimizes either the system message or model weights on the synthetic task (with regularization), then transfers the learned system message or model weights to realistic tasks. We detail each step below. 

\paragraph{Optimization.} 
To reduce the hallucination rate on the synthetic task, \ours{} either optimizes system message $\phi$ or model weights $\theta$. 

\textit{Optimizing the system message} 
$\phi$ is challenging since the set of possible system messages is discrete, precluding continuous optimization with gradient-descent. To circumvent the discreteness of the system messages, we adapt the prefix-tuning method from~\citep{li2021prefix} to learn a continuous postfix to the system message.  
Specifically, we exploit the fact that while prompts are discrete, the LLM maps them to sequences of continuous embeddings during inference. 
To optimize the system message, we append a continuous postfix to the embedded discrete system message, then optimize this postfix with gradient descent. 
Intuitively, the optimized system message provides ``high-level instructions'' that we expect to extrapolate better.  
See \refapp{extra-prefix} for details. 

\textit{Optimizing the model weights} $\theta$ is easy since they are continuous; we use standard fine-tuning.

\paragraph{Regularizing with reference data.} 
Reducing the hallucination rate on the synthetic task may transfer well out-of-the-box, but could potentially pick up on task-specific spurious attributes---attributes that only appear in hallucinated outputs in the synthetic task, but can appear in correct outputs in general. 
For example, models may only output newlines when they hallucinate on the synthetic task, but frequently output newlines in general on realistic tasks. 
Optimizing on the synthetic task may lead the model to never output a newline, which can compromise performance. 

To mitigate the effect of these spurious attributes, we optimize the model to preserve its outputs over a reference distribution. 
Specifically, for reference distribution $\Dref$ of prompts, original system message $\phiog$, model weights $\thetaog$, and KL-divergence $\ell$, we define the reference loss $\lossref$ as
\begin{align}
    \lossref(\phi, \theta; \Dref, \phiog, \thetaog) \coloneqq \mathbb{E}_{\puser \sim \Dref} \ell(\lm_{\phi, \theta}(\puser), \lm_{\phiog, \thetaog}(\puser)). 
\end{align}
Training on the reference data helps combat spurious attributes that can take new values on that data. While LLMs may hallucinate on the reference data (and potentially learn that the true hallucination features are spurious), we find empirically that this is not an issue (\refsec{sim-task-experiments}). 
Optimizing with reference data resembles the PPO-ptx method from \citet{bai2022helpful}, which mixes in pretraining gradients when optimizing a reward model. 

\paragraph{Full loss function.}
We optimize convex combinations of the two losses, i.e.,
\begin{align}
  \lossopt(\phi, \theta; \tausim, \thetaog, \Dref, \alpha)  \coloneqq \alpha \hr(\phi, \theta; \tausim) + (1 - \alpha) \lossref(\phi, \theta, \Dref; \phiog, \thetaog),
    \label{eqn:full-loss}
\end{align}
where $\alpha \in [0,1]$ is a hyperparameter; setting $\alpha = 1$ optimizes on the synthetic task, while $\alpha = 0$ optimizes to preserve the original model. 
We then use the learned parameters on realistic tasks. 

\section{Evaluating \ours{}}
\label{sec:experiments}
We next present an empirical validation of \ours{}. We describe the setup (\refsec{experimental-setup}), show how \ours{} reduces the hallucination rate on realistic tasks (\refsec{sim-task-experiments}), test output quality and other hallucination metrics (\refsec{ablations}), and isolate the impact of the reference data (\refsec{ref-data}).

\subsection{Setup}
\label{sec:experimental-setup}
\paragraph{LLMs.} We evaluate \ours{} on two LLMs: 13B-parameter Vicuna 1.1~\citep{weilin2023vicuna}, and 13B Orca~\citep{mukherjee2023orca}. Both are fine-tuned from Llama 13B~\citep{touvron2023llama}. 

\paragraph{Optimization details.}  
We optimize the system message using prefix tuning~\citep{li2021prefix}, and the entire LLM with standard fine-tuning. 
When optimizing with reference data (denoted Synth. + Ref), we set the factor in Equation \refeqn{full-loss} to $\alpha = 0.5$, and set $\alpha = 0$ for just the synthetic data. 
We only hyperparameter tune Orca on separate MS MARCO validation data, and reuse the hyperparameters all other Orca tasks and all Vicuna tasks; see \refapp{optimization-appendix} for compute and hyperparameter details.

\subsubsection{Realistic tasks}
\label{sec:general-tasks}
We study how well \ours{} reduces hallucination on three realistic LLM use-cases: search-and-retrieve (MS MARCO), meeting summarization (QMSum), and automated clinical report generation (ACI-Bench). Further details for each dataset and full prompts are in \refapp{dataset-details}. 

\paragraph{Search and retrieve (MS MARCO)}. We first study hallucination in search-and-retrieve applications, where the LLM must generate an answer from retrieved documents. 
We use MS MARCO as a source of examples~\citep{nguyen2016ms}, where the task is to answer real-user Bing queries given 10 retrieved passages. 
For computational tractability, we select 1000 random queries from the MS MARCO validation set that require a long-form response (as labeled in the original dataset). 

\paragraph{Meeting summarization (QMSum)}. We next study hallucination in meeting summarization applications, where LLMs are given a meeting transcript and asked to summarize aspects of it. 
We use the QMSum dataset as a source of examples~\citep{zhong2021qmsum}. QMSum contains meeting transcripts and questions (e.g., ``\emph{What did the group discuss about budget balancing?}''). We filter the QMSum train set for entries that fit in the LLM context window, for a total of 852 examples.

\paragraph{Automated clinical report generation (ACI-Bench)}. Finally, we study hallucination in automated clinical report generation. We use ACI-Bench~\citep{yim2023acibench} as a source of examples, which contains dialogs between doctors and patients. Given a dialog, the task is to generate a clinical report with four specific headings. We use all 207 examples from the published dataset. 

\subsubsection{Synthetic tasks}
\label{sec:synthetic-tasks-def}
\paragraph{Synthetic task (\emph{names retrieval})}. 
We define the names retrieval task, where the LLM needs to retrieve certain names from a given list. 
For example, we might prompt the model with the following:
\begin{userinput}
The following is a list of names

[Names]

List the first 5 names where the first name starts with E in the order that they appear. Include both the first and last name in the response. If there are not 5 names that start with E, return all of the names in the list that start with E in the order that they appear. 
\end{userinput}
We generate a dataset of 100,000 examples and test for hallucination by checking whether any name in the output does not appear in the original list. 
When optimizing to reduce hallucination, to use first-order methods, we optimize the log-likelihood of the unique output allowable by the prompt; in \refsec{sim-task-experiments}, we show that doing so also reduces the rate of hallucinated names even among outputs that are not exact matches. See \refapp{names-appendix} for details on the dataset construction. 

\paragraph{Other synthetic tasks.} Before trying the names synthetic task, we tried reversing the words in a sentence, and ``splicing'' pairs of sentences by alternating words from each. 
We found that while LLMs performed poorly on these, they produced outputs that did not match our intuitive definition of hallucination (e.g., when splicing sentences, LLMs frequently just concatenate the two sentences).  
We did not try other synthetic tasks; there are likely others that lead to greater gains. 

\paragraph{Reference data}. For the reference data $\Dref$, we use SQuAD~\citep{rajpurkar2016squad} as a source of 50000 prompts. 
For each passage, we ask the LLM to respond to the associated query, and for half we ask it to explain its reasoning or think step by step. We include details in \refapp{ref-details}. 

\subsection{Reducing hallucination with \ours{}}
\label{sec:sim-task-experiments}
In this section, we measure whether \ours{} teaches LLMs to hallucinate less. 
We first verify that \ours{} reduces the hallucination rate in-domain on the synthetic task, then measure whether it reduces the hallucination rate out-of-domain on realistic tasks. 

\begin{figure}[t]
  \centering
  \includegraphics[width=0.95\linewidth]{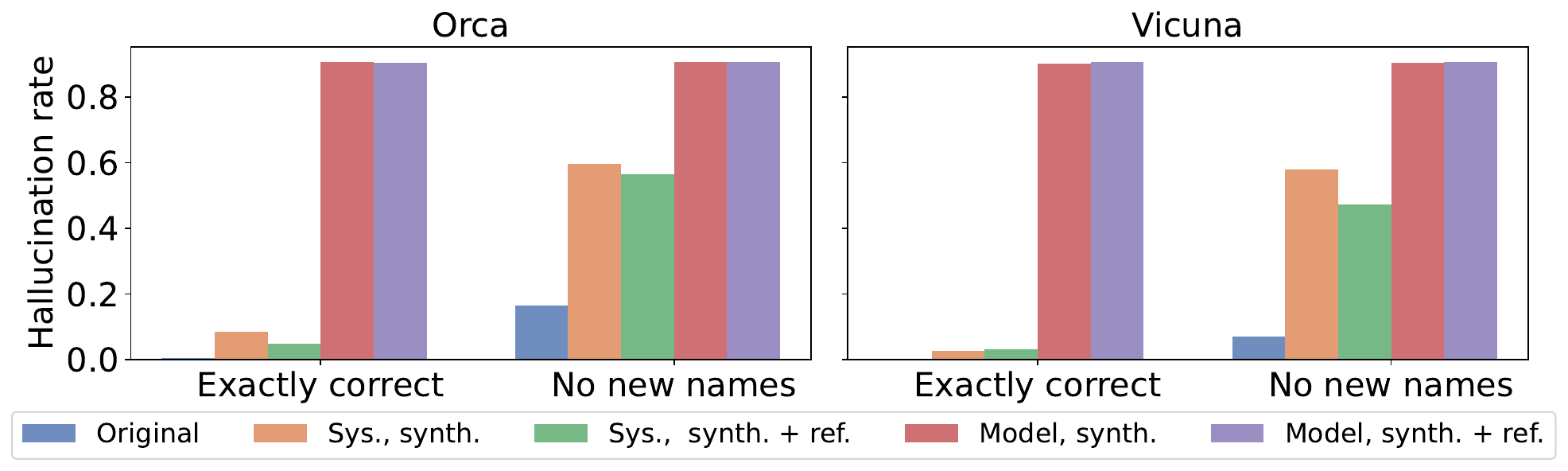}
  \caption{Hallucination rate on the names retrieval task on the original LLM (Original) when optimizing the system message (Sys.) or full LLM weights (Model) on either just the synthetic data (synth.) or mixture of synthetic and reference data (synth. + ref.). We measure whether the LLM is exactly correct, and whether it only generates names on the original list (No new names).}
\label{fig:names_figure} 
\end{figure}
\paragraph{Testing in-domain hallucination reduction.}
We first want to verify that our names retrieval task satisfies our desiderata for a synthetic task, and that our optimizer works. To do so, we want to verify that the unmodified LLM hallucinates frequently on the names retrieval task, and that optimizing on the names retrieval task reduces hallucination on new in-distribution instances. 

In order to test for hallucination on the names retrieval task, we measure whether (i) the generated answer matches the unique correct answer, and (ii) the LLM only generates names on the list.

We report full results in \reffig{names_figure}, and find that the names retrieval task satisfies our desiderata. Both Vicuna and Orca hallucinate frequently on this task; they are exactly correct less than 1\% time, and only generate names on the list less than 17\% of the time. 
Moreover, our optimizer works in-domain; optimizing either the system message or the entire LLM reduces hallucination across all measures.

\paragraph{Testing hallucination reduction on realistic tasks.} We next test whether reducing hallucination on the names retrieval task reduces hallucination on the realistic tasks from \refsec{general-tasks}. We also study how important the reference data and optimization parameters are for transfer performance. 

To measure hallucination on the realistic tasks, we use GPT-4 as an evaluator \citep{openai2023gpt4}, following~\citet{yue2023automatic, gao2023rarr}.\footnote{GPT-4 is tractable to use for our evaluation, but would be expensive to use in optimization directly.}
We prompt GPT-4 with the context, query, and LLM-generated output and ask whether the factual information in the output is grounded up to paraphrases.
GPT-4 retrieves any spans that are not grounded, then scores the groundedness of the response from 0 - 10. We say the LLM hallucinates if the returned score is not 10, i.e., there exists some ungrounded span and thus some fabricated content. See \refapp{gpt4-judge-details} for further details.

\begin{table}
\centering
\begin{tabular}{@{}lllcccc@{}}
\toprule 
Model & Parameters & Data & MS MARCO & QMSum & ACI-Bench & Average\\ 
\midrule
\multirow{5}{*}{Orca} & \multicolumn{2}{c}{\textit{Original model}} & 
$\mathit{12.2 \pm 1.0}$ & $\mathit{18.1 \pm 1.3}$ & $\mathit{47.0 \pm 3.5}$ & $\mathit{25.8}$\\

\cmidrule(lr){2-7} 
 &\multirow{2}{*}{Full model} &Synthetic & $23.7 \pm 1.4$ & $24.8 \pm 1.5$ & $45.4 \pm 3.6$ & $31.3$\\
 & &Synth. + Ref. & $19.6 \pm 1.3$ & $28.0 \pm 1.5$ & $47.0 \pm 3.5$ & $31.5$\\
\cmidrule(lr){2-7} 
 &\multirow{2}{*}{Sys. message} &Synthetic & $\mathbf{9.9 \pm 0.9}$ & $16.2 \pm 1.3$ & $44.8 \pm 3.5$ & $23.6$\\
 & &Synth. + Ref. & $10.5 \pm 1.0$ & $\mathbf{15.8 \pm 1.3}$ & $\mathbf{28.6 \pm 3.3}$ & $\mathbf{18.3}$\\
\midrule\multirow{5}{*}{Vicuna} & \multicolumn{2}{c}{\textit{Original model}} & $\mathit{26.2 \pm 1.4}$ & $\mathit{31.4 \pm 1.6}$ & $\mathit{50.2 \pm 3.5}$ & $\mathit{35.9}$ \\

\cmidrule(lr){2-7} 
 &\multirow{2}{*}{Full model} &Synthetic & $51.3 \pm 1.6$ & $56.9 \pm 1.7$ & $66.3 \pm 3.5$ & $58.2$\\
 & &Synth. + Ref. & $\mathbf{21.1 \pm 1.3}$ & $\mathbf{29.4 \pm 1.6}$ & $50.5 \pm 3.5$ & $33.7$\\
\cmidrule(lr){2-7} 
 &\multirow{2}{*}{Sys. message} &Synthetic & $26.2 \pm 1.4$ & $35.7 \pm 1.6$ & $83.1 \pm 2.6$ & $48.3$\\
 & &Synth. + Ref. & $24.6 \pm 1.4$ & $33.3 \pm 1.6$ & $\mathbf{42.3 \pm 3.5}$ & $\mathbf{33.4}$\\
\bottomrule 
\end{tabular}
\caption{Hallucination rate (\%) measured by GPT-4 across all models, optimized parameters, and tasks (lower is better). We compare against the original model, and optimize the full model or system message, using just synthetic data, or both the synthetic and reference data (\ours{}).}
\label{tab:gpt4-hallucination}
\end{table}

We find that for every realistic task and LLM, \ours{} is able to reduce hallucination by optimizing on a mixture of the names and reference data (\reftab{gpt4-hallucination}). On average, optimizing the system message decreases the hallucination rate by 7.5 points on Orca (a 29\% reduction), and 2.5 points on Vicuna (a 7\% reduction). For specific applications, the decrease can be larger; for example, \ours{} reduces the hallucination rate on ACI-Bench by 16 points on Orca, and 8 points on Vicuna. 

\textbf{Additional analysis.} We next discuss which components are necessary for \ours{} to transfer to realistic tasks, including the optimization parameters, reference, and LLM, in further detail. 

\textit{System message vs LLM weights}. Our results demonstrate that optimizing the system message instead of the whole LLM is sometimes necessary to reduce hallucination, even though fine-tuning is strictly better in-domain. 
On Orca, fine-tuning the full LLM actually \emph{increases} the hallucination rate across most tasks, while optimizing the system message produces consistent reductions. On Vicuna, fine-tuning reduces hallucination on two out of the three tasks, but is slightly worse than optimizing the system message on average. 
We hypothesize that this gap exists in part because fine-tuning latches onto more spurious attributes of the synthetic task, and provide evidence in \refsec{ref-data}. 

\textit{Regularizing with reference data}. Regularizing with reference data is also critical to reduce the hallucination rate; when optimizing the system message, for 5 out of the 6 task-LLM pairs, adding the reference data reduces the hallucination rate. 
While the reference data does not significantly impact the hallucination rate when fine-tuning Orca, it reduces the hallucination rate by over 24 points on average when fine-tuning Vicuna. 
The reference data helps helps break task-specific spurious attributes, which we conjecture is responsible for the improvement (see \refsec{ref-data}). 

\textit{\ours{} reduces the hallucination rate more on Orca than Vicuna.} Finally, \ours{} consistently reduces the hallucination rate more on Orca than Vicuna, and the gains of optimizing the system message instead of the full LLM weights are larger for Orca. 
This is partly because we only hyperparameter tune using Orca; hyperparameter tuning separately on Vicuna was intractable due to computational constraints, but would likely lead to further gains. 
This could also be due to a difference in the quality of the underlying LLMs; the original Orca hallucinates much less than Vicuna, and thus might better leverage optimized instructions, without requiring changes to model weights.

\subsection{Assessing how \ours{} reduces hallucination}
\label{sec:ablations}
We next aim to test \emph{how} \ours{} reduces hallucination, and in particular verify that \ours{} does not simply exploit shortcomings of the GPT-4 evaluation. 
To do so, we test whether the \ours{}-generated outputs are (i) lower-quality and (ii) contain fewer specific details, relative to original outputs.
These tests also provide further evidence that \ours{} reduces the true hallucination rate. 

\paragraph{Testing output quality.} 
We first aim to test whether \ours{} reduces output quality by measuring whether its outputs drift away from reference outputs.
To do so, we compute the BLEU score~\citep{papineni02bleu}, and ROUGE-1, -2, and -L scores~\citep{lin2004rouge}, which compare the $n$-gram overlap between the LLM generated output and the reference output. 
We compute these metrics for MS MARCO and QMSum, as they provide high-quality reference outputs.

\begin{table}
\centering
\begin{adjustbox}{width=.99\textwidth}
\begin{tabular}{@{}lllcccccccc@{}}
\toprule 
Model & Parameters & Data & \multicolumn{4}{c}{MS MARCO} & \multicolumn{4}{c}{QMSum} \\ 
\midrule
 & & & BLEU & R-1 & R-2 & R-L & BLEU & R-1 & R-2 & R-L \\ 
\cmidrule(lr){4-7} \cmidrule(lr){8-11} 
\multirow{5}{*}{Orca} & \multicolumn{2}{c}{\textit{Original model}} & $\mathit{10.5}$ & $\mathit{29.4}$ & $\mathit{18.6}$ & $\mathit{25.3}$ & $\mathit{6.2}$ & $\mathit{34.3}$ & $\mathit{11.1}$ & $\mathit{22.6}$ \\

\cmidrule(lr){2-11} 
 &\multirow{2}{*}{Sys. message} &Synthetic & 14.4 & 35.0 & 23.2 & 30.9 & 6.1 & 34.0 & 11.2 & 22.9\\
 & &Synth. + Ref. & 13.8 & 33.9 & 22.6 & 29.7 & 6.3 & 34.4 & 11.5 & 22.9\\
\cmidrule(lr){2-11} 
 &\multirow{2}{*}{Full model} &Synthetic & 13.9 & 31.6 & 20.4 & 28.8 & 4.0 & 24.0 & 7.8 & 17.1\\
 & &Synth. + Ref. & 11.4 & 30.1 & 19.0 & 26.3 & 5.9 & 33.4 & 11.1 & 22.7\\
\midrule\multirow{5}{*}{Vicuna} & \multicolumn{2}{c}{\textit{Original Model}} & $\mathit{8.7}$ & $\mathit{26.0}$ & $\mathit{15.5}$ & $\mathit{22.1}$ & $\mathit{5.9}$ & $\mathit{33.8}$ & $\mathit{11.0}$ & $\mathit{22.6}$ \\

\cmidrule(lr){2-11} 
 &\multirow{2}{*}{Sys. message} &Synthetic & 8.5 & 26.6 & 15.4 & 22.8 & 5.5 & 33.1 & 10.8 & 22.4\\
 & &Synth. + Ref. & 10.2 & 27.8 & 17.0 & 23.8 & 5.7 & 33.4 & 11.1 & 22.6\\
\cmidrule(lr){2-11} 
 &\multirow{2}{*}{Full model} &Synthetic & 4.9 & 17.7 & 7.9 & 16.3 & 1.8 & 14.8 & 5.0 & 11.4\\
 & &Synth. + Ref. & 11.2 & 29.6 & 18.7 & 25.7 & 5.8 & 33.4 & 10.8 & 22.3\\
\bottomrule 
\end{tabular}
\end{adjustbox}
\caption{{Comparison between outputs generated using \ours{} and human-written reference outputs. We abbreviate ROUGE-1, ROUGE-2, and ROUGE-L with R-1, R-2, and R-L. We find that \ours{} is consistently closer to the reference summaries on MS MARCO than the original model, and is comparable on QMSum. For all metrics, higher is better.}} 
\label{tab:n-gram}
\end{table}

We include full results in \reftab{n-gram}, and find that \ours{}'s outputs do not drift away from reference outputs.
In contrast, when optimizing the system message with reference data, \ours{} has comparable scores across all metrics and LLMs on QMSum, and actually \emph{increases} all metrics on MS MARCO. 
These metrics also reveal that outputs GPT-4 labels as hallucinated are indeed lower quality; Vicuna fine-tuned on only synthetic data, which hallucinates the most, has the lowest scores.  

\textit{What does this say about hallucination?} These metrics provide an indirect signal on whether LLMs hallucinate less; outputs with hallucinated content should be less similar to the fully-grounded reference outputs.\footnote{Since hallucinated content does not appear in reference outputs, hallucination tends to decrease similarity.} 
Thus, \ours{}'s consistent improvement on MS MARCO across all automated metrics for both LLMs provides further evidence that it reduces the true hallucination rate. 

\paragraph{Testing for detail removal.}
We next aim to test whether the LLM avoids hallucinating by generating fewer details. To do so, we use a commercial-grade named entity model that is optimized for healthcare (\refapp{gpt4-judge-details}) to compute all entities in the output and context on ACI-Bench. 
We then test for details by measuring the number of \emph{grounded entities}, i.e., entities in the output that are also in the context. 
To test for hallucination directly, we also measure the number of \emph{ungrounded entities}, i.e., entities in the output that do not appear in the context (and thus could be hallucinated). 
\footnote{An alternative metric would be to test what fraction of outputs have \emph{no} ungrounded entities, but this produces false-positives due noise in the entity model. For example, if ``prescribe'' appears in an output, the entity model extracts ``rib'', in which case the output is considered hallucinated whenever ``rib'' is not in the input.}

\begin{table}
\centering
\begin{adjustbox}{width=.99\textwidth}
\begin{tabular}{@{}lllcccc@{}}
\toprule 
Model & Parameters & Data & Ungrounded $\downbetter$ & Grounded $\upbetter$ &  $\%\!\downarrow$ Ungrounded $\upbetter$ & $\%\!\downarrow$ Grounded $\downbetter$ \\ 
\midrule
\multirow{5}{*}{Orca} & \multicolumn{2}{c}{\textit{Original model}} & $\mathit{9.9}$ & $\mathit{17.2}$ &  -  &  - \\
 &\multirow{2}{*}{Sys. message} &Synthetic & 7.8 & 16.0 & 21.4\% & 6.8\%\\
 & &Synth. + Ref. & 8.0 & 16.9 & 19.2\% & 1.4\%\\
\cmidrule(lr){2-7} 
 &\multirow{2}{*}{Full model} &Synthetic & 2.0 & 5.8 & 79.4\% & 66.2\%\\
 & &Synth. + Ref. & 6.3 & 15.1 & 36.5\% & 12.2\%\\
\midrule
\multirow{5}{*}{Vicuna} & \multicolumn{2}{c}{\textit{Original model}} & $\mathit{9.4}$ & $\mathit{17.6}$ &  -  &  - \\

\cmidrule(lr){2-7} 
 &\multirow{2}{*}{Sys. message} &Synthetic & 4.7 & 11.2 & 49.3\% & 36.5\%\\
 & &Synth. + Ref. & 6.3 & 15.0 & 32.3\% & 14.8\%\\
\cmidrule(lr){2-7} 
 &\multirow{2}{*}{Full model} &Synthetic & 1.4 & 0.2 & 84.9\% & 99.0\%\\
 & &Synth. + Ref. & 5.3 & 14.5 & 43.2\% & 17.6\%\\
\bottomrule 
\end{tabular}
\end{adjustbox}
\caption{Entity evaluation between the prompts and outputs in ACI-Bench. Using a NER model, we measure the number of entities that are in the output that do not appear in the prompt (Ungrounded, lower is better), and that do appear in the prompt (Grounded, higher is better), along with the percentage decrease in ungrounded and grounded entities relative to the original output.} 
\label{tab:entities}
\end{table}
We include full results in \reftab{entities}, and find that \ours{} does not significantly reduce the number of grounded entities; it decreases the number of grounded entities that Orca generates by 1.4\%, and that Vicuna generates by 14.8\%. 
However, \ours{} decreases the number of ungrounded entities by much more: by 19.2\% for Orca and 36.5\% for Vicuna. Fine-tuning and training without the reference data eliminate more grounded and ungrounded entities. 

\textit{What does this say about hallucination?} These results provide a direct signal that \ours{} reduces the hallucination rate by decreasing the number of ungrounded entities across all tested methods.  

\subsection{Reference data combats spurious attributes}
\label{sec:ref-data}
We next aim to identify whether optimizing on the reference data combats task-specific spurious attributes, which can drive up the hallucination rate. 
To do so, we identify newlines as an easy-to-evaluate spurious attribute.
In the names retrieval task, all correct answers do not have newlines, so LLMs may learn that newlines are associated with hallucinations. 
However, never outputting a newline could lead to errors when transferring; for example, LLMs may abruptly end generation before answering the query, rather than outputting a newline (e.g., ``\emph{The items are:}'' [ends]). 

We report the newline rate across all methods and tasks in \reftab{newline} of \refapp{additional-reference}, and find that training on reference data helps mitigate the effect of the spurious attributes.  
Training on the names retrieval task routinely reduces the newline rate, but adding in the reference data recovers much of the drop. 
We measure the presence of newlines as we identified it as a potential spurious attribute and could measure it easily, but reference data likely similarly combats unknown spurious attributes. 

\section{Related Work}
\paragraph{Hallucination.} We aim to reduce hallucination in text generation systems when all salient information is included in context. Text generation systems frequently hallucinate; see a general survey~\citep{ji2023survey}, and surveys restricted to abstractive summarization~\citep{maynez2020faithfulness, huang2021factual} for examples. Hallucination is one of many documented potential risks of deploying LLMs~\citep{bender2021stochastic, bommasani2021opportunities, weidinger2021ethical}. 

Hallucination is hard to detect automatically.  
Some work measures hallucination by comparing outputs to reference summaries using BLEU score~\citep{papineni02bleu}, ROUGE score~\citep{lin2004rouge}, or entity overlap \citep{nan2021entity}.
More recent work measures hallucination by decomposing outputs into atomic facts and evaluating them~\citep{min2023factscore}, or uses LLMs~\citep{yue2023automatic, gao2023rarr}. Another line of work suggests that LLMs may encode whether they are hallucinating within their activations~\citep{kadavath2022language, burns2023discovering, azaria2023internal}, but this has not been scaled to abstractive summarization settings. 

There are a few classes of methods to reduce hallucination. Some methods adjust the generation process, either by changing the decoding strategy~\citep{tian2019sticking, shi2023trusting}, teaching the LLM to cite \citep{gao2023enabling}, edit~\citep{gao2023rarr}, or abstain~\citep{cao2023learning} as it generates, or incorporating knowledge graphs~\citep{ji2023rho} and external documents~\citep{mallen2023trust}.
One line of work aims to reduce hallucination with prompting strategies~\citep{jung2022maieutic, zhou2023context}, while another edits model internals to make LLMs more honest~\citep{li2023inference}, or faithful to the context~\citep{hernandez2023inspecting}. 
The closest line of work to ours trains models to hallucinate less with contrastive learning~\citep{cao2021cliff, tang2022confit} or reinforcement learning~\citep{roit2023factually}. 
Our work optimizes to reduce hallucination directly on synthetic data.

Other work optimizes for other LLM behaviors such as helpfulness or harmlessness. They do so by learning a reward function capturing the behavior with human feedback~\citep{sadigh2017active, christiano2017deep}, then optimizing LLMs using this reward function~\citep{steinnon2020learning, bai2022helpful, ouyang2022instructions}. 
Such work aims to learn an approximate a general-purpose objective to optimize, while \ours{} optimizes an exact objective on a narrower domain. 

\paragraph{Synthetic data.} \ours{} leverages synthetic data to reduce hallucination. 
Synthetic data generated by LLMs has been used to train higher-quality small models~\citep{eldan2023tinystories, gunasekar2023textbooks}, train models to follow instructions~\citep{dubois2023alpacafarm, weilin2023vicuna, mukherjee2023orca}, and make models better at clinical text mining~\citep{tang2023synthetic}. 
A closer line of work to ours trains LLMs with synthetic data that comes from non-LLM sources;~\citet{sanh2021multitask} convert existing NLP benchmarks to tasks for instruction tuning,~\citet{wei2023symbol} add random synthetic labels, and~\citet{wei2023simple} adapt benchmarks to reduce sycophancy. 
The closest work to ours is~\citet{zhang2023hallucinations}, which aims to characterize hallucination by evaluating it on synthetic tasks. 

\paragraph{Prompt optimization.} \ours{} reduces hallucination by optimizing the LLM system message; to do so we append a continuous postfix, then optimize using prefix-tuning~\citep{li2021prefix}. 
Both~\citet{li2021prefix} and~\citet{su2022transferability} demonstrate that prefixes can transfer well between classification tasks; our work provides further evidence of this for generative tasks. 
Some work aims to optimize a discrete prompt directly to improve performance on classification tasks~\citep{shin2020autoprompt, wen2023hard}; such methods could in principle be plugged into \ours{} directly. 
\section{Discussion}
\label{sec:discussion}
We introduce \ours{} a method to reduce hallucination by defining and exploiting synthetic tasks. 
\ours{} reduces the hallucination rate across a suite of realistic evaluation tasks.

There are many natural ways to improve \ours{}. We could improve the optimization method by searching for better hyperparameters for fine-tuning and prefix-tuning, testing other fine-tuning methods like LoRA~\citep{hu2022lora}, or jointly optimizing the system prompt and model weights (rather than each in isolation). 
We could adapt \ours{} to reduce hallucination on models that are accessible only via APIs, and scale \ours{} to larger open-source models. 
And finally, we could extract more from synthetic tasks, by (i) searching for a synthetic task that reduces the hallucination rate more over all tasks, (ii) constructing separate synthetic tasks for each realistic task, or (iii) training on multiple synthetic tasks jointly. We expect many of these would further reduce the hallucination rate. 

While we give heuristics for choosing synthetic tasks, we do not know what properties causally reduce hallucination.  
The mechanism driving hallucinations in the names retrieval task likely overlaps with that of the real tasks; this makes the learned system message and models transfer to realistic tasks.
However, pinning down properties explicitly would help us define synthetic tasks more systematically, and determine which other LLM behaviors (such as generating unsafe of unhelpful outputs) can be tamed with synthetic data; this is an exciting direction for subsequent work.

Finally, our work poses a natural question: \emph{When is defining a synthetic task better than collecting human feedback on realistic tasks directly?}
Synthetic tasks allow us to scalably generate lots of data to optimize on, and makes biases clearer. However, it can be hard to identify good synthetic tasks, and methods like \ours{} require a loss-inducing transfer step after optimization. 
In contrast, optimizing on demonstrations eliminates the transfer step and may generalize more broadly, but requires expensive-to-obtain feedback that can bias the resulting model in unexpected ways \citep{caspser2023rlhf}. 
Understanding when collecting synthetic data is more effective than demonstrations could help better allocate resources as we aim ensure that LLMs are reliable, helpful, and safe.

\subsubsection*{Acknowledgments}
We thank Besmira Nushi, Alex Pan, Mert Yuksekgonul, Olivia Watkins, Jacob Steinhardt, and Ruiqi Zhong for feedback on this work. 

\bibliography{all.bib}

\begin{thebibliography}{63}
\providecommand{\natexlab}[1]{#1}
\providecommand{\url}[1]{\texttt{#1}}
\expandafter\ifx\csname urlstyle\endcsname\relax
  \providecommand{\doi}[1]{doi: #1}\else
  \providecommand{\doi}{doi: \begingroup \urlstyle{rm}\Url}\fi

\bibitem[Azaria \& Mitchell(2023)Azaria and Mitchell]{azaria2023internal}
Amos Azaria and Tom Mitchell.
\newblock The internal state of an {LLM} knows when its lying.
\newblock \emph{arXiv preprint arXiv:2304.13734}, 2023.

\bibitem[Bai et~al.(2022)Bai, Jones, Ndousse, Askell, Chen, DasSarma, Drain,
  Fort, Ganguli, Henighan, Joseph, Kadavath, Kernion, Conerly, El-Showk,
  Elhage, Hatfield-Dodds, Hernandez, Hume, Johnston, Kravec, Lovitt, Nanda,
  Olsson, Amodei, Brown, Clark, McCandlish, Olah, Mann, and
  Kaplan]{bai2022helpful}
Yuntao Bai, Andy Jones, Kamal Ndousse, Amanda Askell, Anna Chen, Nova DasSarma,
  Dawn Drain, Stanislav Fort, Deep Ganguli, T.~Henighan, Nicholas Joseph,
  Saurav Kadavath, John Kernion, Tom Conerly, S.~El-Showk, Nelson Elhage, Zac
  Hatfield-Dodds, Danny Hernandez, Tristan Hume, Scott Johnston, S.~Kravec,
  Liane Lovitt, Neel Nanda, Catherine Olsson, Dario Amodei, Tom~B. Brown, Jack
  Clark, Sam McCandlish, C.~Olah, Benjamin Mann, and J.~Kaplan.
\newblock Training a helpful and harmless assistant with reinforcement learning
  from human feedback.
\newblock \emph{arXiv}, 2022.

\bibitem[Bender et~al.(2021)Bender, Gebru, McMillan-Major, and
  Shmitchel]{bender2021stochastic}
Emily Bender, Timnit Gebru, Angelina McMillan-Major, and Shmargaret Shmitchel.
\newblock On the dangers of stochastic parrots: Can language models be too big?
\newblock In \emph{ACM Conference on Fairness, Accountability, and Transparency
  (FAccT)}, 2021.

\bibitem[Bommasani et~al.(2021)Bommasani, Hudson, Adeli, Altman, Arora, von
  Arx, Bernstein, Bohg, Bosselut, Brunskill, Brynjolfsson, Buch, Card,
  Castellon, Chatterji, Chen, Creel, Davis, Demszky, Donahue, Doumbouya,
  Durmus, Ermon, Etchemendy, Ethayarajh, Fei-Fei, Finn, Gale, Gillespie, Goel,
  Goodman, Grossman, Guha, Hashimoto, Henderson, Hewitt, Ho, Hong, Hsu, Huang,
  Icard, Jain, Jurafsky, Kalluri, Karamcheti, Keeling, Khani, Khattab, Koh,
  Krass, Krishna, Kuditipudi, Kumar, Ladhak, Lee, Lee, Leskovec, Levent, Li,
  Li, Ma, Malik, Manning, Mirchandani, Mitchell, Munyikwa, Nair, Narayan,
  Narayanan, Newman, Nie, Niebles, Nilforoshan, Nyarko, Ogut, Orr,
  Papadimitriou, Park, Piech, Portelance, Potts, Raghunathan, Reich, Ren, Rong,
  Roohani, Ruiz, Ryan, Ré, Sadigh, Sagawa, Santhanam, Shih, Srinivasan,
  Tamkin, Taori, Thomas, Tramèr, Wang, Wang, Wu, Wu, Wu, Xie, Yasunaga, You,
  Zaharia, Zhang, Zhang, Zhang, Zhang, Zheng, Zhou, and
  Liang]{bommasani2021opportunities}
Rishi Bommasani, Drew~A. Hudson, Ehsan Adeli, Russ Altman, Simran Arora, Sydney
  von Arx, Michael~S. Bernstein, Jeannette Bohg, Antoine Bosselut, Emma
  Brunskill, Erik Brynjolfsson, Shyamal Buch, Dallas Card, Rodrigo Castellon,
  Niladri Chatterji, Annie Chen, Kathleen Creel, Jared~Quincy Davis, Dorottya
  Demszky, Chris Donahue, Moussa Doumbouya, Esin Durmus, Stefano Ermon, John
  Etchemendy, Kawin Ethayarajh, Li~Fei-Fei, Chelsea Finn, Trevor Gale, Lauren
  Gillespie, Karan Goel, Noah Goodman, Shelby Grossman, Neel Guha, Tatsunori
  Hashimoto, Peter Henderson, John Hewitt, Daniel~E. Ho, Jenny Hong, Kyle Hsu,
  Jing Huang, Thomas Icard, Saahil Jain, Dan Jurafsky, Pratyusha Kalluri,
  Siddharth Karamcheti, Geoff Keeling, Fereshte Khani, Omar Khattab, Pang~Wei
  Koh, Mark Krass, Ranjay Krishna, Rohith Kuditipudi, Ananya Kumar, Faisal
  Ladhak, Mina Lee, Tony Lee, Jure Leskovec, Isabelle Levent, Xiang~Lisa Li,
  Xuechen Li, Tengyu Ma, Ali Malik, Christopher~D. Manning, Suvir Mirchandani,
  Eric Mitchell, Zanele Munyikwa, Suraj Nair, Avanika Narayan, Deepak
  Narayanan, Ben Newman, Allen Nie, Juan~Carlos Niebles, Hamed Nilforoshan,
  Julian Nyarko, Giray Ogut, Laurel Orr, Isabel Papadimitriou, Joon~Sung Park,
  Chris Piech, Eva Portelance, Christopher Potts, Aditi Raghunathan, Rob Reich,
  Hongyu Ren, Frieda Rong, Yusuf Roohani, Camilo Ruiz, Jack Ryan, Christopher
  Ré, Dorsa Sadigh, Shiori Sagawa, Keshav Santhanam, Andy Shih, Krishnan
  Srinivasan, Alex Tamkin, Rohan Taori, Armin~W. Thomas, Florian Tramèr,
  Rose~E. Wang, William Wang, Bohan Wu, Jiajun Wu, Yuhuai Wu, Sang~Michael Xie,
  Michihiro Yasunaga, Jiaxuan You, Matei Zaharia, Michael Zhang, Tianyi Zhang,
  Xikun Zhang, Yuhui Zhang, Lucia Zheng, Kaitlyn Zhou, and Percy Liang.
\newblock On the opportunities and risks of foundation models.
\newblock \emph{arXiv preprint arXiv:2108.07258}, 2021.

\bibitem[Burns et~al.(2023)Burns, Ye, Klein, and
  Steinhardt]{burns2023discovering}
Collin Burns, Haotian Ye, Dan Klein, and Jacob Steinhardt.
\newblock Discovering latent knowledge in language models without supervision.
\newblock In \emph{International Conference on Learning Representations
  (ICLR)}, 2023.

\bibitem[Cao et~al.(2023)Cao, Dong, He, and Cheung]{cao2023learning}
Meng Cao, Yue Dong, Jingyi He, and Jackie Chi~Kit Cheung.
\newblock Learning with rejection for abstractive text summarization.
\newblock \emph{arXiv preprint arXiv:2302.08531}, 2023.

\bibitem[Cao \& Wang(2021)Cao and Wang]{cao2021cliff}
Shuyang Cao and Lu~Wang.
\newblock {CLIFF}: Contrastive learning for improving faithfulness and
  factuality in abstractive summarization.
\newblock In \emph{Empirical Methods in Natural Language Processing (EMNLP)},
  2021.

\bibitem[Casper et~al.(2023)Casper, Davies, Shi, Gilbert, Scheurer, Rando,
  Freedman, Korbak, Lindner, Freire, Wang, Marks, Segerie, Carroll, Peng,
  Christoffersen, Damani, Slocum, Anwar, Siththaranjan, Nadeau, Michaud, Pfau,
  Krasheninnikov, Chen, Langosco, Hase, Bıyık, Dragan, Krueger, Sadigh, and
  Hadfield-Menell]{caspser2023rlhf}
Stephen Casper, Xander Davies, Claudia Shi, Thomas~Krendl Gilbert, Jérémy
  Scheurer, Javier Rando, Rachel Freedman, Tomasz Korbak, David Lindner, Pedro
  Freire, Tony Wang, Samuel Marks, Charbel-Raphaël Segerie, Micah Carroll,
  Andi Peng, Phillip Christoffersen, Mehul Damani, Stewart Slocum, Usman Anwar,
  Anand Siththaranjan, Max Nadeau, Eric~J. Michaud, Jacob Pfau, Dmitrii
  Krasheninnikov, Xin Chen, Lauro Langosco, Peter Hase, Erdem Bıyık, Anca
  Dragan, David Krueger, Dorsa Sadigh, and Dylan Hadfield-Menell.
\newblock Open problems and fundamental limitations of reinforcement learning
  from human feedback.
\newblock \emph{arXiv preprint arXiv:2307.15217}, 2023.

\bibitem[Chiang et~al.(2023)Chiang, Li, Lin, Sheng, Wu, Zhang, Zheng, Zhuang,
  Zhuang, Gonzalez, Stoica, and Xing]{weilin2023vicuna}
Wei-Lin Chiang, Zhuohan Li, Zi~Lin, Ying Sheng, Zhanghao Wu, Hao Zhang, Lianmin
  Zheng, Siyuan Zhuang, Yonghao Zhuang, Joseph~E. Gonzalez, Ion Stoica, and
  Eric~P. Xing.
\newblock Vicuna: An open-source chatbot impressing {GPT-4} with 90
  quality.
\newblock https://lmsys.org/blog/2023-03-30-vicuna/, 2023.

\bibitem[Christiano et~al.(2017)Christiano, Leike, Brown, Martic, Legg, and
  Amodei]{christiano2017deep}
Paul Christiano, Jan Leike, Tom~B. Brown, Miljan Martic, Shane Legg, and Dario
  Amodei.
\newblock Deep reinforcement learning from human preferences.
\newblock In \emph{Advances in Neural Information Processing Systems
  (NeurIPS)}, 2017.

\bibitem[Dubois et~al.(2023)Dubois, Li, Taori, Zhang, Gulrajani, Ba, Guestrin,
  Liang, and Hashimoto]{dubois2023alpacafarm}
Yann Dubois, Xuechen Li, Rohan Taori, Tianyi Zhang, Ishaan Gulrajani, Jimmy Ba,
  Carlos Guestrin, Percy Liang, and Tatsunori~B. Hashimoto.
\newblock {AlpacaFarm}: A simulation framework for methods that learn from
  human feedback.
\newblock \emph{arXiv preprint arXiv:2305.14387}, 2023.

\bibitem[Eldan \& Li(2023)Eldan and Li]{eldan2023tinystories}
Ronen Eldan and Yuanzhi Li.
\newblock Tinystories: How small can language models be and still speak
  coherent english?
\newblock \emph{arXiv preprint arXiv:2305.00759}, 2023.

\bibitem[Gao et~al.(2023{\natexlab{a}})Gao, Dai, Pasupat, Chen, Chaganty, Fan,
  Zhao, Lao, Lee, Juan, and Guu]{gao2023rarr}
Luyu Gao, Zhuyun Dai, Panupong Pasupat, Anthony Chen, Arun~Tejasvi Chaganty,
  Yicheng Fan, Vincent~Y. Zhao, Ni~Lao, Hongrae Lee, Da-Cheng Juan, and Kelvin
  Guu.
\newblock Rarr: Researching and revising what language models say, using
  language models.
\newblock In \emph{Association for Computational Linguistics (ACL)},
  2023{\natexlab{a}}.

\bibitem[Gao et~al.(2023{\natexlab{b}})Gao, Yen, Yu, and Chen]{gao2023enabling}
Tianyu Gao, Howard Yen, Jiatong Yu, and Danqi Chen.
\newblock Enabling large language models to generate text with citations.
\newblock \emph{arXiv preprint arXiv:2305.14627}, 2023{\natexlab{b}}.

\bibitem[Guerreiro et~al.(2023)Guerreiro, Voita, and
  Martins]{guerriero2023looking}
Nuno~M. Guerreiro, Elena Voita, and André Martins.
\newblock Looking for a needle in a haystack: A comprehensive study of
  hallucinations in neural machine translation.
\newblock In \emph{European Association for Computational Linguistics (EACL)},
  2023.

\bibitem[Gunasekar et~al.(2023)Gunasekar, Zhang, Aneja, Mendes, Giorno, Gopi,
  Javaheripi, Kauffmann, de~Rosa, Saarikivi, Salim, Shah, Behl, Wang, Bubeck,
  Eldan, Kalai, Lee, and Li]{gunasekar2023textbooks}
Suriya Gunasekar, Yi~Zhang, Jyoti Aneja, Caio César~Teodoro Mendes, Allie~Del
  Giorno, Sivakanth Gopi, Mojan Javaheripi, Piero Kauffmann, Gustavo de~Rosa,
  Olli Saarikivi, Adil Salim, Shital Shah, Harkirat~Singh Behl, Xin Wang,
  Sébastien Bubeck, Ronen Eldan, Adam~Tauman Kalai, Yin~Tat Lee, and Yuanzhi
  Li.
\newblock Textbooks are all you need.
\newblock \emph{arXiv preprint arXiv:2306.11644}, 2023.

\bibitem[Hernandez et~al.(2023)Hernandez, Li, and
  Andreas]{hernandez2023inspecting}
Evan Hernandez, Belinda~Z. Li, and Jacob Andreas.
\newblock Inspecting and editing knowledge representations in language models.
\newblock \emph{arXiv preprint arXiv:2304.00740}, 2023.

\bibitem[Hu et~al.(2022)Hu, Shen, Wallis, Allen-Zhu, Li, Wang, Wang, and
  Chen]{hu2022lora}
Edward~J. Hu, Yelong Shen, Phillip Wallis, Zeyuan Allen-Zhu, Yuanzhi Li, Shean
  Wang, Lu~Wang, and Weizhu Chen.
\newblock Lora: Low-rank adaptation of large language models.
\newblock In \emph{International Conference on Learning Representations
  (ICLR)}, 2022.

\bibitem[Huang et~al.(2021)Huang, Feng, Feng, and Qin]{huang2021factual}
Yichong Huang, Xiachong Feng, Xiaocheng Feng, and Bing Qin.
\newblock The factual inconsistency problem in abstractive text summarization:
  A survey.
\newblock \emph{arXiv preprint arXiv:2104.14839}, 2021.

\bibitem[Ji et~al.(2023{\natexlab{a}})Ji, Lee, Frieske, Yu, Su, Xu, Ishii,
  Bang, Dai, Madotto, and Fung]{ji2023survey}
Ziwei Ji, Nayeon Lee, Rita Frieske, Tiezheng Yu, Dan Su, Yan Xu, Etsuko Ishii,
  Yejin Bang, Wenliang Dai, Andrea Madotto, and Pascale Fung.
\newblock Survey of hallucination in natural language generation.
\newblock \emph{ACM Computing Surveys (CSUR)}, 55, 2023{\natexlab{a}}.

\bibitem[Ji et~al.(2023{\natexlab{b}})Ji, Liu, Lee, Yu, Wilie, Zeng, and
  Fung]{ji2023rho}
Ziwei Ji, Zihan Liu, Nayeon Lee, Tiezheng Yu, Bryan Wilie, Min Zeng, and
  Pascale Fung.
\newblock {RHO}: Reducing hallucination in open-domain dialogues with knowledge
  grounding.
\newblock In \emph{Findings of the Association for Computational Linguistics
  (Findings of ACL)}, 2023{\natexlab{b}}.

\bibitem[Jung et~al.(2022)Jung, Qin, Welleck, Brahman, Bhagavatula, Bras, and
  Choi]{jung2022maieutic}
Jaehun Jung, Lianhui Qin, Sean Welleck, Faeze Brahman, Chandra Bhagavatula,
  Ronan~Le Bras, and Yejin Choi.
\newblock Maieutic prompting: Logically consistent reasoning with recursive
  explanations.
\newblock In \emph{Empirical Methods in Natural Language Processing (EMNLP)},
  2022.

\bibitem[Kadavath et~al.(2022)Kadavath, Conerly, Askell, Henighan, Drain,
  Perez, Schiefer, Hatfield-Dodds, DasSarma, Tran-Johnson, Johnston, El-Showk,
  Jones, Elhage, Hume, Chen, Bai, Bowman, Fort, Ganguli, Hernandez, Jacobson,
  Kernion, Kravec, Lovitt, Ndousse, Olsson, Ringer, Amodei, Brown, Clark,
  Joseph, Mann, McCandlish, Olah, and Kaplan]{kadavath2022language}
Saurav Kadavath, Tom Conerly, Amanda Askell, Tom Henighan, Dawn Drain, Ethan
  Perez, Nicholas Schiefer, Zac Hatfield-Dodds, Nova DasSarma, Eli
  Tran-Johnson, Scott Johnston, Sheer El-Showk, Andy Jones, Nelson Elhage,
  Tristan Hume, Anna Chen, Yuntao Bai, Sam Bowman, Stanislav Fort, Deep
  Ganguli, Danny Hernandez, Josh Jacobson, Jackson Kernion, Shauna Kravec,
  Liane Lovitt, Kamal Ndousse, Catherine Olsson, Sam Ringer, Dario Amodei, Tom
  Brown, Jack Clark, Nicholas Joseph, Ben Mann, Sam McCandlish, Chris Olah, and
  Jared Kaplan.
\newblock Language models (mostly) know what they know.
\newblock \emph{arXiv preprint arXiv:2207.05221}, 2022.

\bibitem[Kingma \& Ba(2015)Kingma and Ba]{kingma2015adam}
Diederik Kingma and Jimmy Ba.
\newblock Adam: A method for stochastic optimization.
\newblock In \emph{International Conference on Learning Representations
  (ICLR)}, 2015.

\bibitem[Koto et~al.(2022)Koto, Lau, and Baldwin]{koto2022pretrained}
Fajri Koto, Jey~Han Lau, and Timothy Baldwin.
\newblock Can pretrained language models generate persuasive, faithful, and
  informative ad text for product descriptions?
\newblock In \emph{Empirical Methods in Natural Language Processing (EMNLP)},
  2022.

\bibitem[Li et~al.(2023)Li, Patel, Viégas, Pfister, and
  Wattenberg]{li2023inference}
Kenneth Li, Oam Patel, Fernanda Viégas, Hanspeter Pfister, and Martin
  Wattenberg.
\newblock Inference-time intervention: Eliciting truthful answers from a
  language model.
\newblock In \emph{Advances in Neural Information Processing Systems
  (NeurIPS)}, 2023.

\bibitem[Li \& Liang(2021)Li and Liang]{li2021prefix}
Xiang~Lisa Li and Percy Liang.
\newblock Prefix-tuning: Optimizing continuous prompts for generation.
\newblock In \emph{Association for Computational Linguistics (ACL)}, 2021.

\bibitem[Liu et~al.(2023)Liu, Zhang, and Liang]{liu2023evaluating}
Nelson~F. Liu, Tianyi Zhang, and Percy Liang.
\newblock Evaluating verifiability in generative search engines.
\newblock \emph{arXiv}, 2023.

\bibitem[Mallen et~al.(2023)Mallen, Asai, Zhong, Das, Khashabi, and
  Hajishirzi]{mallen2023trust}
Alex Mallen, Akari Asai, Victor Zhong, Rajarshi Das, Daniel Khashabi, and
  Hannaneh Hajishirzi.
\newblock When not to trust language models: Investigating effectiveness of
  parametric and non-parametric memories.
\newblock In \emph{Association for Computational Linguistics (ACL)}, 2023.

\bibitem[Maynez et~al.(2020)Maynez, Narayan, Bohnet, and
  McDonald]{maynez2020faithfulness}
Joshua Maynez, Shashi Narayan, Bernd Bohnet, and Ryan McDonald.
\newblock On faithfulness and factuality in abstractive summarization.
\newblock In \emph{Association for Computational Linguistics (ACL)}, 2020.

\bibitem[Min et~al.(2023)Min, Krishna, Lyu, Lewis, tau Yih, Koh, Iyyer,
  Zettlemoyer, and Hajishirzi]{min2023factscore}
Sewon Min, Kalpesh Krishna, Xinxi Lyu, Mike Lewis, Wen tau Yih, Pang~Wei Koh,
  Mohit Iyyer, Luke Zettlemoyer, and Hannaneh Hajishirzi.
\newblock {FActScore}: Fine-grained atomic evaluation of factual precision in
  long form text generation.
\newblock \emph{arXiv preprint arXiv:2305.14251}, 2023.

\bibitem[Mukherjee et~al.(2023)Mukherjee, Mitra, Jawahar, Agarwal, Palangi, and
  Awadallah]{mukherjee2023orca}
Subhabrata Mukherjee, Arindam Mitra, Ganesh Jawahar, Sahaj Agarwal, Hamid
  Palangi, and Ahmed Awadallah.
\newblock Orca: Progressive learning from complex explanation traces of
  {GPT-4}.
\newblock \emph{arXiv preprint arXiv:2306.02707}, 2023.

\bibitem[Nan et~al.(2021)Nan, Nallapati, Wang, dos Santos, Zhu, Zhang, Mckeown,
  and Xiang]{nan2021entity}
Feng Nan, Ramesh Nallapati, Zhiguo Wang, Cicero dos Santos, Henghui Zhu, Dejiao
  Zhang, Kathleen Mckeown, and Bing Xiang.
\newblock Entity-level factual consistency of abstractive text summarization.
\newblock In \emph{European Association for Computational Linguistics (EACL)},
  2021.

\bibitem[Nguyen et~al.(2016)Nguyen, Rosenberg, Song, Gao, Tiwary, Majumder, and
  Deng]{nguyen2016ms}
Tri Nguyen, Mir Rosenberg, Xia Song, Jianfeng Gao, Saurabh Tiwary, Rangan
  Majumder, and Li~Deng.
\newblock {MS MARCO}: A human generated machine reading comprehension dataset.
\newblock In \emph{Workshop on Cognitive Computing at NIPS}, 2016.

\bibitem[{OpenAI}(2023)]{openai2023gpt4}
{OpenAI}.
\newblock {GPT}-4 technical report.
\newblock \emph{arXiv preprint arXiv:2303.08774}, 2023.

\bibitem[Ouyang et~al.(2022)Ouyang, Wu, Jiang, Almeida, Wainwright, Mishkin,
  Zhang, Agarwal, Slama, Ray, Schulman, Hilton, Kelton, Miller, Simens, Askell,
  Welinder, Christiano, Leike, and Lowe]{ouyang2022instructions}
Long Ouyang, Jeff Wu, Xu~Jiang, Diogo Almeida, Carroll~L. Wainwright, Pamela
  Mishkin, Chong Zhang, Sandhini Agarwal, Katarina Slama, Alex Ray,
  J.~Schulman, Jacob Hilton, Fraser Kelton, Luke~E. Miller, Maddie Simens,
  Amanda Askell, P.~Welinder, P.~Christiano, J.~Leike, and Ryan~J. Lowe.
\newblock Training language models to follow instructions with human feedback.
\newblock \emph{arXiv}, 2022.

\bibitem[Papineni et~al.(2002)Papineni, Roukos, Ward, and Zhu]{papineni02bleu}
Kishore Papineni, Salim Roukos, Todd Ward, and Wei-Jing Zhu.
\newblock {BLEU}: A method for automatic evaluation of machine translation.
\newblock In \emph{Association for Computational Linguistics (ACL)}, 2002.

\bibitem[Post(2018)]{post2018call}
Matt Post.
\newblock A call for clarity in reporting {BLEU} scores.
\newblock In \emph{Proceedings of the Third Conference on Machine Translation:
  Research Papers}, 2018.

\bibitem[Rajpurkar et~al.(2016)Rajpurkar, Zhang, Lopyrev, and
  Liang]{rajpurkar2016squad}
Pranav Rajpurkar, Jian Zhang, Konstantin Lopyrev, and Percy Liang.
\newblock {SQuAD}: 100,000+ questions for machine comprehension of text.
\newblock In \emph{Empirical Methods in Natural Language Processing (EMNLP)},
  2016.

\bibitem[Remy(2021)]{remy2021names}
Philippe Remy.
\newblock Names dataset.
\newblock https://github.com/philipperemy/name-dataset, 2021.

\bibitem[Roit et~al.(2023)Roit, Ferret, Shani, Aharoni, Cideron, Dadashi,
  Geist, Girgin, Hussenot, Keller, Momchev, Ramos, Stanczyk, Vieillard, Bachem,
  Elidan, Hassidim, Pietquin, and Szpektor]{roit2023factually}
Paul Roit, Johan Ferret, Lior Shani, Roee Aharoni, Geoffrey Cideron, Robert
  Dadashi, Matthieu Geist, Sertan Girgin, Léonard Hussenot, Orgad Keller,
  Nikola Momchev, Sabela Ramos, Piotr Stanczyk, Nino Vieillard, Olivier Bachem,
  Gal Elidan, Avinatan Hassidim, Olivier Pietquin, and Idan Szpektor.
\newblock Factually consistent summarization via reinforcement learning with
  textual entailment feedback.
\newblock In \emph{Association for Computational Linguistics (ACL)}, 2023.

\bibitem[Sadigh et~al.(2017)Sadigh, Dragan, Sastry, and
  Seshia]{sadigh2017active}
Dorsa Sadigh, Anca Dragan, Shankar Sastry, and Sanjit Seshia.
\newblock Active preference-based learning of reward functions.
\newblock In \emph{Robotics: Science and Systems (RSS)}, 2017.

\bibitem[Sanh et~al.(2021)Sanh, Webson, Raffel, Bach, Sutawika, Alyafeai,
  Chaffin, Stiegler, Scao, Raja, Dey, Bari, Xu, Thakker, Sharma, Szczechla,
  Kim, Chhablani, Nayak, Datta, Chang, Jiang, Wang, Manica, Shen, Yong, Pandey,
  Bawden, Wang, Neeraj, Rozen, Sharma, Santilli, Fevry, Fries, Teehan,
  Biderman, Gao, Bers, Wolf, and Rush]{sanh2021multitask}
Victor Sanh, Albert Webson, Colin Raffel, Stephen~H. Bach, Lintang Sutawika,
  Zaid Alyafeai, Antoine Chaffin, Arnaud Stiegler, Teven~Le Scao, Arun Raja,
  Manan Dey, M~Saiful Bari, Canwen Xu, Urmish Thakker, Shanya~Sharma Sharma,
  Eliza Szczechla, Taewoon Kim, Gunjan Chhablani, Nihal Nayak, Debajyoti Datta,
  Jonathan Chang, Mike Tian-Jian Jiang, Han Wang, Matteo Manica, Sheng Shen,
  Zheng~Xin Yong, Harshit Pandey, Rachel Bawden, Thomas Wang, Trishala Neeraj,
  Jos Rozen, Abheesht Sharma, Andrea Santilli, Thibault Fevry, Jason~Alan
  Fries, Ryan Teehan, Stella Biderman, Leo Gao, Tali Bers, Thomas Wolf, and
  Alexander~M. Rush.
\newblock Multitask prompted training enables zero-shot task generalization.
\newblock \emph{arXiv}, 2021.

\bibitem[Shi et~al.(2023)Shi, Han, Lewis, Tsvetkov, Zettlemoyer, and tau
  Yih]{shi2023trusting}
Weijia Shi, Xiaochuang Han, Mike Lewis, Yulia Tsvetkov, Luke Zettlemoyer, and
  Scott~Wen tau Yih.
\newblock Trusting your evidence: Hallucinate less with context-aware decoding.
\newblock \emph{arXiv preprint arXiv:2305.14739}, 2023.

\bibitem[Shin et~al.(2020)Shin, Razeghi, IV, Wallace, and
  Singh]{shin2020autoprompt}
Taylor Shin, Yasaman Razeghi, Robert L.~Logan IV, Eric Wallace, and Sameer
  Singh.
\newblock Autoprompt: Eliciting knowledge from language models with
  automatically generated prompts.
\newblock In \emph{Empirical Methods in Natural Language Processing (EMNLP)},
  2020.

\bibitem[Stiennon et~al.(2020)Stiennon, Ouyang, Wu, Ziegler, Lowe, Voss,
  Radford, Amodei, and Christiano]{steinnon2020learning}
Nisan Stiennon, Long Ouyang, Jeff Wu, Daniel~M. Ziegler, Ryan Lowe, Chelsea
  Voss, Alec Radford, Dario Amodei, and Paul Christiano.
\newblock Learning to summarize from human feedback.
\newblock In \emph{Advances in Neural Information Processing Systems
  (NeurIPS)}, 2020.

\bibitem[Su et~al.(2022)Su, Wang, Qin, Chan, Lin, Wang, Wen, Liu, Li, Li, Hou,
  Sun, and Zhou]{su2022transferability}
Yusheng Su, Xiaozhi Wang, Yujia Qin, Chi-Min Chan, Yankai Lin, Huadong Wang,
  Kaiyue Wen, Zhiyuan Liu, Peng Li, Juanzi Li, Lei Hou, Maosong Sun, and Jie
  Zhou.
\newblock On transferability of prompt tuning for natural language processing.
\newblock In \emph{North American Association for Computational Linguistics
  (NAACL)}, 2022.

\bibitem[Tang et~al.(2023)Tang, Han, Jiang, and Hu]{tang2023synthetic}
Ruixiang Tang, Xiaotian Han, Xiaoqian Jiang, and Xia Hu.
\newblock Does synthetic data generation of llms help clinical text mining?
\newblock \emph{arXiv preprint arXiv:2303/04360}, 2023.

\bibitem[Tang et~al.(2022)Tang, Nair, Wang, Wang, Desai, Wade, Li, Celikyilmaz,
  Mehdad, and Radev]{tang2022confit}
Xiangru Tang, Arjun Nair, Borui Wang, Bingyao Wang, Jai Desai, Aaron Wade,
  Haoran Li, Asli Celikyilmaz, Yashar Mehdad, and Dragomir Radev.
\newblock Confit: Toward faithful dialogue summarization with
  linguistically-informed contrastive fine-tuning.
\newblock In \emph{North American Association for Computational Linguistics
  (NAACL)}, 2022.

\bibitem[Tian et~al.(2019)Tian, Narayan, Sellam, and Parikh]{tian2019sticking}
Ran Tian, Shashi Narayan, Thibault Sellam, and Ankur~P. Parikh.
\newblock Sticking to the facts: Confident decoding for faithful data-to-text
  generation.
\newblock \emph{arXiv preprint arXiv:1910.08684}, 2019.

\bibitem[Touvron et~al.(2023)Touvron, Lavril, Izacard, Martinet, Lachaux,
  Lacroix, Rozière, Goyal, Hambro, Azhar, Rodriguez, Joulin, Grave, and
  Lample]{touvron2023llama}
Hugo Touvron, Thibaut Lavril, Gautier Izacard, Xavier Martinet, Marie-Anne
  Lachaux, Timothée Lacroix, Baptiste Rozière, Naman Goyal, Eric Hambro,
  Faisal Azhar, Aurelien Rodriguez, Armand Joulin, Edouard Grave, and Guillaume
  Lample.
\newblock Llama: Open and efficient foundation language models.
\newblock \emph{arXiv}, 2023.

\bibitem[Wei et~al.(2022)Wei, Wang, Schuurmans, Bosma, Ichter, Xia, Chi, Le,
  and Zhou]{wei2022cot}
Jason Wei, Xuezhi Wang, Dale Schuurmans, Maarten Bosma, Brian Ichter, Fei Xia,
  Ed~Chi, Quoc Le, and Denny Zhou.
\newblock Chain-of-thought prompting elicits reasoning in large language
  models.
\newblock \emph{arXiv preprint arXiv:2201.11903}, 2022.

\bibitem[Wei et~al.(2023{\natexlab{a}})Wei, Hou, Lampinen, Chen, Huang, Tay,
  Chen, Lu, Zhou, Ma, and Le]{wei2023symbol}
Jerry Wei, Le~Hou, Andrew Lampinen, Xiangning Chen, Da~Huang, Yi~Tay, Xinyun
  Chen, Yifeng Lu, Denny Zhou, Tengyu Ma, and Quoc~V. Le.
\newblock Symbol tuning improves in-context learning in language models.
\newblock \emph{arXiv preprint arXiv:2305.08298}, 2023{\natexlab{a}}.

\bibitem[Wei et~al.(2023{\natexlab{b}})Wei, Huang, Lu, Zhou, and
  Le]{wei2023simple}
Jerry Wei, Da~Huang, Yifeng Lu, Denny Zhou, and Quoc~V. Le.
\newblock Simple synthetic data reduces sycophancy in large language models.
\newblock \emph{arXiv preprint arXiv:2308.03958}, 2023{\natexlab{b}}.

\bibitem[Weidinger et~al.(2021)Weidinger, Mellor, Rauh, Griffin, Uesato, Huang,
  Cheng, Glaese, Balle, Kasirzadeh, Kenton, Brown, Hawkins, Stepleton, Biles,
  Birhane, Haas, Rimell, Hendricks, Isaac, Legassick, Irving, and
  Gabriel]{weidinger2021ethical}
Laura Weidinger, John Mellor, Maribeth Rauh, Conor Griffin, Jonathan Uesato,
  Po-Sen Huang, Myra Cheng, Mia Glaese, Borja Balle, Atoosa Kasirzadeh, Zac
  Kenton, Sasha Brown, Will Hawkins, Tom Stepleton, Courtney Biles, Abeba
  Birhane, Julia Haas, Laura Rimell, Lisa~Anne Hendricks, William Isaac, Sean
  Legassick, Geoffrey Irving, and Iason Gabriel.
\newblock Ethical and social risks of harm from language models.
\newblock \emph{arXiv preprint arXiv:2112.04359}, 2021.

\bibitem[Wen et~al.(2023)Wen, Jain, Kirchenbauer, Goldblum, Geiping, and
  Goldstein]{wen2023hard}
Yuxin Wen, Neel Jain, John Kirchenbauer, Micah Goldblum, Jonas Geiping, and Tom
  Goldstein.
\newblock Hard prompts made easy: Gradient-based discrete optimization for
  prompt tuning and discovery.
\newblock \emph{arXiv preprint arXiv:2302.03668}, 2023.

\bibitem[Wolf et~al.(2019)Wolf, Debut, Sanh, Chaumond, Delangue, Moi, Cistac,
  Rault, Louf, Funtowicz, and Brew]{wolf2019transformers}
Thomas Wolf, Lysandre Debut, Victor Sanh, Julien Chaumond, Clement Delangue,
  Anthony Moi, Pierric Cistac, Tim Rault, R'emi Louf, Morgan Funtowicz, and
  Jamie Brew.
\newblock {HuggingFace}'s transformers: State-of-the-art natural language
  processing.
\newblock \emph{arXiv preprint arXiv:1910.03771}, 2019.

\bibitem[yew Lin \& Rey(2004)yew Lin and Rey]{lin2004rouge}
Chin yew Lin and Marina Rey.
\newblock Looking for a few good metrics: {ROUGE} and its evaluation.
\newblock In \emph{NTCIR Workshop}, 2004.

\bibitem[Yim et~al.(2023)Yim, Fu, Abacha, Snider, Lin, and
  Yetisgen]{yim2023acibench}
Wenwai Yim, Yujuan Fu, Asma~Ben Abacha, Neal Snider, Thomas Lin, and Meliha
  Yetisgen.
\newblock {ACI-BENCH}: a novel ambient clinical intelligence dataset for
  benchmarking automatic visit note generation.
\newblock \emph{arXiv preprint arXiv:2306.02022}, 2023.

\bibitem[Yue et~al.(2023)Yue, Wang, Zhang, Chen, Su, and Sun]{yue2023automatic}
Xiang Yue, Boshi Wang, Kai Zhang, Ziru Chen, Yu~Su, and Huan Sun.
\newblock Automatic evaluation of attribution by large language model.
\newblock \emph{arXiv preprint arXiv:2305.06311}, 2023.

\bibitem[Zhang et~al.(2023)Zhang, Press, Merrill, Liu, and
  Smith]{zhang2023hallucinations}
Muru Zhang, Ofir Press, William Merrill, Alisa Liu, and Noah~A. Smith.
\newblock How language model hallucinations can snowball.
\newblock \emph{arXiv preprint arXiv:2305.13534}, 2023.

\bibitem[Zhong et~al.(2021)Zhong, Yin, Yu, Zaidi, Mutuma, Jha, Awadallah,
  Celikyilmaz, Liu, Qiu, and Radev]{zhong2021qmsum}
Ming Zhong, Da~Yin, Tao Yu, Ahmad Zaidi, Mutethia Mutuma, Rahul Jha,
  Ahmed~Hassan Awadallah, Asli Celikyilmaz, Yang Liu, Xipeng Qiu, and Dragomir
  Radev.
\newblock Qmsum: A new benchmark for query-based multi-domain meeting
  summarization.
\newblock In \emph{North American Association for Computational Linguistics
  (NAACL)}, 2021.

\bibitem[Zhou et~al.(2023)Zhou, Zhang, Poon, and Chen]{zhou2023context}
Wenxuan Zhou, Sheng Zhang, Hoifung Poon, and Muhao Chen.
\newblock Context-faithful prompting for large language models.
\newblock \emph{arXiv preprint arXiv:2303.11315}, 2023.

\end{thebibliography}
\bibliographystyle{iclr2024_conference}

\newpage
\appendix
\section{Additional details for The \ours{} pipeline}
\label{sec:extra-prefix}
In this section, we provide additional details on our adaption of \citet{li2021prefix}'s prefix-tuning method. 

\paragraph{LLM preprocessing.} We first define some prelimaries for LLM preprocessing.
Given a prompt $p$, the language model splits the prompt into tokens with the tokenizer function $t$, i.e., $t(p) = p_1, \hdots, p_m$, where $p_i$ is the $i^{th}$ token of the tokenized prompt. 
Language models then map each token to a continuous embedding using a simple lookup table, and concatenate them. This produces an embedded prompt $e(p)$ that is a $d \times m$ matrix where $d$ is the dimension of the language model embeddings. 

\paragraph{Appending the postfix}. To add the postfix to the system message, we simply initialize a random $d \times n$ matrix, then append it to the embedded system message to form a $d \times n + m$ matrix. 

In practice, the implementations of Orca and Vicuna embed the system message and prompt jointly. To identify where the system prompt ends, we exploit the formatting they use and insert the system prompt right before the first ``newline'' embedding, ensuring it is placed correctly. 

\paragraph{Optimizing the postfix.} To optimize the postfix, \citet{li2021prefix} introduce a lower-dimensional vector that they optimize, then transform back to embedding space with a matrix. 
Instead, we optimize the postfix in the full embedding dimension, which is 5120 for both variants of Llama that we consider. 
\citet{li2021prefix} do this to stabilize optimization for smaller models; our work suggests that this might not be necessary for larger models. 
\section{Additional details for Evaluating \ours{}}
\label{sec:additional-experimental-details}

\subsection{Optimization, inference, and hyperparameter details.}
\label{sec:optimization-appendix}
In this section, we describe the specific experimental details used. 

\paragraph{Compute details.} We perform all of our experiments on a single NVIDIA A100-PCIE-80GB GPU, except for fine-tuning, for which we use four A100s. 
We run inference and optimization for all models in bfloat16 precision. 
We use 13B parameter models, as these were the largest that could fit in memory when optimizing the postfix. 

\paragraph{Inference details.}
For all experiments (including the original LLMs with no optimization), we use the system message  ``\emph{You are a helpful, honest, and conservative AI system designed to answer queries using only the provided context.}''. Intuitively, we aim to make the baseline LLMs hallucinate less via prompting, so that the gains made by optimizing do not rediscover easy prompting gains. 
We sample with temperature 0.7 when generating, and have a max sequence length of 1024 tokens. Since the context length of the models we study is 2048 tokens, prompts that have more than 1024 tokens can only have up to 2048 combined tokens in the prompt and output. 

\paragraph{Prefix tuning details.}
We append a $n = 10$ token postfix to the embedded system message, which has 51200 parameters (10 tokens times model embedding dimension 5120). 
We initialize the postfix to be 10 copies of Llama's ``space'' token embedding. 
We optimize the postfix with Adam \citep{kingma2015adam}, using learning rate 1e-4, no weight decay, epsilon 1e-7 and otherwise the default HuggingFace parameters \citep{wolf2019transformers}. 
We chose the learning rate out of two options (1e-3 and 1e-4), and the number of tokens out of three options (1, 10, and 20) based on the hallucination rate using Orca on a separate MS-Marco validation set; we then recycle the same hyperparameters for all other realistic tasks and models. This means that \ours{} is not hyperparameter tuned on Vicuna at all. 
We use episilon 1e-7 to avoid numerical stability issues that arise when converting models to bfloat16. 

\paragraph{Fine-tuning details}
In order to fine-tune the baseline models Orca and Vicuna 1.1, both of which have 13 billion parameters, we use copy the hyperparameters from \citep{weilin2023vicuna}. Specifically, we use a learning rate of 5e-5, warm up ratio of 0.03, weight decay is 0, and we run fine-tuning for one epoch with batch size of 12 per device on four NVIDIA A100-PCIE-80GB GPUs in bfloat16 precision. 

For every method, we optimize for one epoch which is exactly 100000 examples. When optimizing on synthetic data mixed with the reference data, we use 50000 examples on the names task and 50000 examples on the reference task. 

\subsection{Dataset details}
\label{sec:dataset-details}
We next describe the dataset details for each of the realistic tasks we study. 

\textbf{MS MARCO.} 
We first describe how we get our MS MARCO subset, using the data from \citep{nguyen2016ms}. 
We start with the validation set, then 
filter for examples where response type is a "description" (i.e., requires a long-form response), and where at least one document is useful for producing the answer (to enforce that the model only needs to use the context to answer). These labels are included in the data. 
Of the remaining examples, we randomly choose 1000. 

For MS Marco, we use the following prompt
\begin{userinput}
\#\#\# System: You are a helpful, honest, and conservative AI system designed to answer queries using only the provided context.

\#\#\# Human: The following is a list of passages:

-[Passage 1]

...

-[Passage 10]

Using the passages, respond to the following query:

[query]

\#\#\# Assistant:
\end{userinput}
Where here the passages and the query are from the dataset. 

\textbf{QMSum.} 
We next curate a subset of QMSum \citep{zhong2021qmsum}. To do so, we start with the training set, then take all queries from the "specific query list" associated with each meeting. For each query in the list, we filter the meeting transcript based on the "relevant text spans" as our transcript, then only include those that have under 1800 tokens according to the Llama tokenizer (so models can generate at least 248 tokens in response). 
This produces 852 examples. We use the training set rather than the validation because the validation set has far fewer examples, and we don't optimize against this dataset directly. 

For QMSum we use the following prompt
\begin{userinput}
\#\#\# System: You are a helpful, honest, and conservative AI system designed to answer queries using only the provided context.

\#\#\# Human: The following is a meeting transcript:

[relevant lines of meeting transcript]

Using the transcript, respond to the following query:

Query: [query]

\#\#\# Assistant:
\end{userinput}
Where the relevant lines of the meeting transcript and the query are taken directly from the dataset. 

\textbf{ACI-Bench.} 
Finally, we describe our adaptation of the ACI-Bench dataset \citep{yim2023acibench}. 
We combine the train, validation, and three test splits for a total of 207 examples; this is all the possible examples in the dataset. Each examples contains a dialog between a patient and doctor. 

We use the prompt that \citet{yim2023acibench} use with our system prompt directly. This gives us the following prompt: 
\begin{userinput}
\#\#\# System: You are a helpful, honest, and conservative AI system designed to answer queries using only the provided context.

\#\#\# Human: Summarize the conversation to generate a clinical note with four sections: HISTORY OF PRESENT ILLNESS, PHYSICAL EXAM, RESULTS, ASSESSMENT AND PLAN. The conversation is: [dialog]. 

\#\#\# Assistant:
\end{userinput}
Here, the dialogues are taken directly from the dataset. 

\subsection{Names data}
\label{sec:names-appendix}
We next describe our names synthetic data. 
We generate names by taking the top 50000 first and last names in the U.S. from \citet{remy2021names}, then from these select 100 random first and last names, then combine them. 

\begin{userinput}
\#\#\# System: You are a helpful, honest, and conservative AI system designed to answer queries using only the provided context.

\#\#\# Human: The following is a list of names

[Name 1]

...

[Name 100]

List the first 5 names where the first name starts with [first letter] in the order that they appear. Include both the first and last name in the response. If there are not 5 names that start with [first letter], return all of the names in the list that start with [first letter] in the order that they appear. 

\#\#\# Assistant: 
\end{userinput}
Here, the first letter is randomly chosen among the all letters for which there is at least one name, and the names are randomly generated according to the above procedure. 

In \reffig{names_figure}, we evaluate on 1000 randomly generated names prompts that are separate from the 100000 training examples. 

\subsection{Reference data}
\label{sec:ref-details}
As a source of reference data, we use SQuAD \citep{rajpurkar2016squad}. 
Specifically, we take random SQuAD passages and their associated queries to create the following prompts.
\begin{userinput}
\#\#\# System: You are a helpful, honest, and conservative AI system designed to answer queries using only the provided context.

\#\#\# Human: The following is a passage:

[SQuAD passage]

Using the passage, respond to the following query

[query]

[response type]

\#\#\# Assistant:
\end{userinput}
Here, SQuAD passage and query come directly from the SQuAD dataset. To get a more diverse set of prompts and outputs, we try different options for [response type]. For 25000 examples we leave it empty, and for 25000 examples we use chain-of-thought style prompts \citep{wei2022cot}; for 10000 examples set response type to "While performing the task explain your reasoning", and for 15000 examples we set the response type to a chain-of-thought style prompt: "While performing the task think step-by-step and justify your steps". 

\subsection{Additional evaluation details}
\label{sec:gpt4-judge-details}

\paragraph{GPT-4 evaluation.} In this section, we describe how we used GPT-4 as a judge, following \citet{yue2023automatic}. 
The prompt in \citet{yue2023automatic} asks GPT-4 if the context supports / contradicts the response, or if there's not enough information. 
However, we found that using this prompt reported hallucination rates of under 3\%, which we empirically found was far lower than the true hallucination rate. 

Instead, we prompt GPT-4 to score whether each piece of information in the model response is supported by the context, and return ungrounded spans. We encourage it to check each piece of information from the reply, especially focusing on numbers, dates, names, etc., without worrying about paraphrases. We describe the task of GPT-4 as validating the outputs of another language model. 
The full prompt is proprietary, and mimics prompts used to test for grounding in production. 

\paragraph{Similarity evaluation.} We next provide additional details for computing the BLUE, ROUGE-1, -2, -and -L scores from figure \reftab{n-gram}. When computing BLEU score, we use the method introduced in \citep{post2018call}. When there are multiple possible reference summaries available, we choose the last reference summary. 

\paragraph{Entity evaluation.} For retrieval of healthcare entities on the ACI-Bench dataset, we used a Named-Entity Recognition service from Microsoft Azure called \textbf{Text Analytics for Health}\footnote{Azure Text Analytics for
health is one of the pre-built features offered by Azure AI Language Service. It is a cloud-based API service
that applies machine-learning intelligence to extract and label relevant
medical information from a variety of unstructured texts such as doctor's
notes, discharge summaries, clinical documents, and electronic health records. More information available on: https://learn.microsoft.com/en-us/azure/ai-services/language-service/text-analytics-for-health/overview?tabs=ner}.  
In order to reduce noise, we select a subset of medical and regular entity types to be extracted from the documents, and a minimum confidence score of 0.75.  
The healthcare entities that we select include \textit{Allergen, BodyStructure, ConditionQualifier, ConditionScale, Course,  Diagnosis, Direction, ExaminationName, Expression, GeneOrProtein,
MedicationClass, MedicationForm, MedicationName, MedicationRoute,
MutationType, SubstanceUse, SymptomOrSign, TreatmentName} and \textit{Variant}.  
We include yet one more non-healthcare entity, \textit{Quantity}, as numerical reasoning is known to be a challenging task that can lead to hallucination more easily \citep{ji2023survey}. 

\subsection{Additional reference data results}
\label{sec:additional-reference}
\begin{table}
\centering
\begin{tabular}{@{}lllcccc@{}}
\toprule 
Model & Parameters & Data & MS MARCO & QMSum & ACI-bench & Average\\ 
\midrule
\multirow{5}{*}{Orca} 
& \multicolumn{2}{c}{\textit{Original model}} & $\mathit{12.4}$ & $\mathit{4.7}$ & $\mathit{88.4}$ & $\mathit{35.2}$ \\

\cmidrule(lr){2-7} 
 &\multirow{2}{*}{Sys. message} &Synthetic & $6.7$ & $2.0$ & $86.5$ & $31.7$\\
 & &Synth. + Ref. & $6.9$ & $3.8$ & $88.4$ & $33.0$\\
\cmidrule(lr){2-7} 
 &\multirow{2}{*}{Full model} &Synthetic & $0.0$ & $0.2$ & $16.9$ & $5.7$\\
 & &Synth. + Ref. & $4.7$ & $3.2$ & $77.8$ & $28.5$\\
\midrule\multirow{5}{*}{Vicuna} 
& \multicolumn{2}{c}{\textit{Original model}} & $\mathit{6.9}$ & $\mathit{2.2}$ & $\mathit{83.1}$ & $\mathit{30.7}$ \\

\cmidrule(lr){2-7} 
 &\multirow{2}{*}{Sys. message} &Synthetic & $2.8$ & $0.7$ & $28.5$ & $10.7$\\
 & &Synth. + Ref. & $3.8$ & $0.7$ & $57.0$ & $20.5$\\
\cmidrule(lr){2-7} 
 &\multirow{2}{*}{Full model} &Synthetic & $0.0$ & $0.4$ & $22.7$ & $7.7$\\
 & &Synth. + Ref. & $4.3$ & $2.7$ & $82.6$ & $29.9$\\
\bottomrule 
\end{tabular}
\caption{Newline rate (\%) when optimizing either the system message or full model on just synthetic or synthetic + reference data, for both Orca and Vicuna. Optimizing on just the synthetic data tends to reduce the newline rate (a spurious attribute), adding the reference data restores it.} 
\label{tab:newline}
\end{table}
In this section, we present the newline experiment results. We include the results in \reftab{newline}, and find that the newline rate tends to decrease on the synthetic data, but increases again when training on the synthetic and reference data jointly. The newline rate is a proxy for other spurious attributes that are harder to anticipate and measure. 
\end{document}